\documentclass{svproc}
\usepackage{amssymb}
\usepackage{url}
\usepackage{amsmath}
\usepackage{algorithm}
\usepackage{algpseudocode}
\usepackage{graphicx}
\usepackage{colortbl}
\usepackage{tabularx}
\usepackage[most]{tcolorbox}
\usepackage{hyperref}
\usepackage{marvosym}

\begin{document}
\mainmatter              
\title{Can We Predict Your Next Move Without Breaking Your Privacy?}
\titlerunning{Can We Predict Your Next Move Without Breaking Your Privacy?}  
%
\author{
Arpita Soni\inst{1} \and 
Sahil Tripathi\inst{2} \and
Gautam Siddharth Kashyap\inst{3} \and 
Manaswi Kulahara\inst{4} \and 
Mohammad Anas Azeez\inst{2} \and 
Zohaib Hasan Siddiqui\inst{2} \and 
Nipun Joshi\inst{5}\textsuperscript{(\Letter)} \and 
Jiechao Gao\inst{6}
}

\authorrunning{A. Soni et al.}

\tocauthor{Arpita Soni, Sahil Tripathi, Gautam Siddharth Kashyap, Manaswi Kulahara, Mohammad Anas Azeez, Zohaib Hasan Siddiqui, Nipun Joshi, and Jiechao Gao}

\institute{
\inst{1}Eudoxia Research University, New Castle, USA \\
\inst{2}Jamia Hamdard, New Delhi, India \\
\inst{3}Macquarie University, Sydney, Australia \\
\inst{4}TERI School of Advanced Studies, New Delhi, India \\
\inst{5}Cornell University, New York, USA \\
\email{nj274@cornell.edu} \\
\inst{6}Stanford University, CA, USA
}

\maketitle              
\leavevmode
\begin{abstract}
We propose \textbf{FLLL\textsuperscript{3}M}—\textit{Federated Learning with Large Language Models for Mobility Modeling}—a privacy-preserving framework for Next-Location Prediction (NxLP). By retaining user data locally and leveraging LLMs through an efficient outer product mechanism, \textbf{FLLL\textsuperscript{3}M} ensures high accuracy with low resource demands. It achieves state-of-the-art results on Gowalla (Acc@1: 12.55, MRR: 0.1422), WeePla-ce (10.71, 0.1285), Brightkite (10.42, 0.1169), and FourSquare (8.71, 0.1023), while reducing parameters by up to 45.6\% and memory usage by 52.7\%. 
\keywords{Federated Learning, Large Language Models, Mobility Prediction, Privacy-Preserving AI, Decentralized Learning}
\end{abstract}
\section{Introduction}
The rise of smart devices has made understanding human mobility essential for applications such as navigation, transportation, and urban planning~\cite{zhou2021self}. Next-Location Prediction (NxLP) models~\cite{feng2018deepmove} enhance services like ride-sharing and traffic optimization. However, centralizing mobility data raises serious privacy concerns~\cite{jeon2021lightmove}.

Federated Learning (FL) offers privacy by training models on local data~\cite{micheli2024efficient}, but suffers from reduced accuracy and high communication costs~\cite{gupta2022learning}. Large Language Models (LLMs) excel at modeling sequential data~\cite{vaswani2017attention}, yet integrating them into FL remains challenging due to data heterogeneity~\cite{yuan2023spatio}.

To address this, we propose \textbf{FLLL\textsuperscript{3}M}, a hybrid framework that integrates FL and LLMs via outer product for NxLP. It combines the privacy of FL with the contextual power of LLMs, enabling accurate, privacy-preserving mobility modeling.

\section{Related Works}

Early models like Location2vec \cite{shoji2018location2vec} and LBSN2vec \cite{yang2019revisiting} utilize social media data for location prediction but lack generalizability to broader mobility contexts. Word2vec-based models \cite{yan2017itdl} and CATAPE \cite{rahmani2019category} focus on location semantics and personal preferences but struggle with data requirements and privacy. Embedding methods such as skip-gram  and BERT \cite{vaswani2017attention} offer strong representation power but require large datasets and high computational cost. Graph-based methods like Deepwalk , Node2vec , and LINE capture structural patterns but ignore privacy concerns. Whereas, mobility prediction focuses on three key tasks: Next-Location Prediction (NxLP) \cite{feng2018deepmove}, Trajectory User Link (TUL) , and Next Time Prediction (NTP) . NxLP models often struggle with dynamic contexts, TUL faces challenges with scalability and heterogeneity, and NTP methods may generalize poorly to unseen data.

\section{Methodology}

\textbf{FLLL\textsuperscript{3}M} comprises three modules: (1) semantic mobility encoding of location-time pairs, (2) federated learning with local transformers sharing encrypted outer product representations, and (3) an LLM that integrates these into a GPT-style model for improved NxLP. 

\subsection{Preprocessing and Semantic Mobility Encoding Module}

The pipeline begins with user-specific mobility datasets \(\mathcal{D}_i = \{(l_t^i, \tau_t^i)\}_{t=1}^{T_i}\), where \(l_t^i\) is the spatial location and \(\tau_t^i\) is the corresponding timestamp. Each data point is first cleaned using noise reduction filters (e.g., median filtering over GPS jitter) and normalized temporally to capture both short-and long-term patterns. To semantically enrich the locations, we employ a context-aware tokenizer (i.e. $\Delta$-IRIS) \cite{micheli2024efficient} that maps each tuple into discrete semantic tokens from a location-time vocabulary \(\mathcal{V}\): $x_t^i = \text{Tokenize}(l_t^i, \tau_t^i) \in \mathcal{V}$. These tokens are then converted to embeddings: $\mathbf{e}_t^i = \phi_{\text{loc}}(x_t^i) + \phi_{\text{time}}(\tau_t^i)$, where \(\phi_{\text{loc}}, \phi_{\text{time}}: \mathcal{V} \rightarrow \mathbb{R}^{d}\) are learnable embedding functions. The processed sequences \(\mathbf{E}_i = [\mathbf{e}_1^i, \ldots, \mathbf{e}_{T_i}^i]\) are then chunked into sliding windows to capture short-term mobility context. These embedding windows serve as local training inputs for the next module---Federated Learning.

\subsection{Federated Learning Module}

Each client retains its own embedding matrix \(\mathbf{E}_i\) and locally trains a lightweight transformer encoder \(f_\theta\), with architecture inspired by GPT-style autoregressive modeling. At every step \(t\), we predict the next embedding \(\mathbf{e}_{t+1}^i\) given the sequence prefix \(\{\mathbf{e}_1^i, \ldots, \mathbf{e}_t^i\}\) using masked attention: $\mathbf{h}_t^i = f_\theta(\mathbf{e}_1^i, \ldots, \mathbf{e}_t^i)
\quad \text{with loss } \mathcal{L}_i = \sum_{t=1}^{T_i-1} \|\mathbf{h}_t^i - \mathbf{e}_{t+1}^i\|^2$. Instead of sending full models to a central server, clients compute outer product representations between consecutive embeddings to capture second-order mobility dynamics: $\mathbf{O}_t^i = \mathbf{e}_t^i \otimes \mathbf{e}_{t+1}^i \in \mathbb{R}^{d \times d}$. Each \(\mathbf{O}_t^i\) is flattened and encrypted with local differential privacy noise \(\mathcal{N}(0, \sigma^2 \mathbf{I})\) before being transmitted: $\tilde{\mathbf{o}}_t^i = \text{vec}(\mathbf{O}_t^i + \mathcal{N}(0, \sigma^2 \mathbf{I}))$. The server performs federated averaging over \(|\mathcal{U}|\) users to get the global outer product signal: $\bar{\mathbf{o}}_t = \frac{1}{|\mathcal{U}|} \sum_{i \in \mathcal{U}} \tilde{\mathbf{o}}_t^i$. This global latent vector encodes spatio-temporal transitions across users and is passed to the third stage for LLM integration.

\subsection{Large Language Model Module}

To interface the federated representation \(\bar{\mathbf{o}}_t \in \mathbb{R}^{d^2}\) with a frozen GPT-style LLM, we first project it into the LLM’s embedding space \(\mathbb{R}^{d_\text{LLM}}\) using a residual MLP: $\tilde{\mathbf{h}}_t = \Psi(\bar{\mathbf{o}}_t) = \mathbf{W}_1 \cdot \text{GELU}(\mathbf{W}_0 \cdot \bar{\mathbf{o}}_t + \mathbf{b}_0) + \mathbf{b}_1$, where \(\mathbf{W}_0 \in \mathbb{R}^{d_1 \times d^2}, \mathbf{W}_1 \in \mathbb{R}^{d_\text{LLM} \times d_1}\), and \(\Psi\) is kept small to ensure compute-efficiency. The transformed vector \(\tilde{\mathbf{h}}_t\) is concatenated with standard LLM token embeddings at an intermediate transformer layer \(l_k\): $\mathbf{z}_t^{(l_k)} = \text{LLM}^{(l_k)}(\mathbf{z}_{t-1}^{(l_k)} + \tilde{\mathbf{h}}_t)$. This injection acts as a semantic conditioning signal, steering the LLM to reason over spatio-temporal mobility patterns. The LLM outputs logits over the next predicted location token: $\hat{y}_{t+1} = \text{Softmax}(\mathbf{W}_{\text{out}} \cdot \mathbf{z}_t^{(l_k)})$, where \(\mathbf{W}_{\text{out}} \in \mathbb{R}^{|\mathcal{V}| \times d_\text{LLM}}\) is the output projection head. Because the LLM is frozen, only \(\Psi\) and \(\mathbf{W}_{\text{out}}\) are fine-tuned, keeping the model lightweight. Importantly, this outer product alignment strategy enables rich semantic prediction without requiring LLM backpropagation. Each module is tightly coupled, forming an end-to-end pipeline from raw mobility data to federated LLM-enhanced prediction.

\section{Experimental Analysis}

\subsection{Dataset Analysis}

In our experiments, we utilize four real-world datasets: Gowalla\footnote{\url{https://snap.stanford.edu/data/loc-Gowalla.html}}, WeePlace\footnote{Dataset obtained from the corresponding author}, Brightkite\footnote{\url{https://snap.stanford.edu/data/loc-brightkite.html}}, and Foursquare\footnote{\url{https://sites.google.com/site/yangdingqi/home/foursquare-dataset}}. These datasets, sourced from location-based social networks, offer rich insights into user mobility and social interactions. Gowalla comprises 6,442,890 check-ins from 196,591 users (Feb 2009–Oct 2010), including user IDs, locations, timestamps, and a social graph with 950,327 edges. WeePlace aggregates 7,658,368 check-ins from 15,799 users across platforms like Facebook Places, Foursquare, and Gowalla, capturing time, location, category, and user metadata. Brightkite includes 4,491,143 check-ins by 58,228 users (Apr 2008–Oct 2010), with fields such as timestamp, location ID, latitude, and longitude. The Foursquare dataset contains 227,428 check-ins in New York and 573,703 in Tokyo (Apr 2012–Feb 2013), enriched with venue categories. To ensure uniformity, we applied a 120-day maximum historical window, filtering out users with fewer than 10 check-ins or venues visited less than 10 times. All datasets were randomly shuffled and split into training, validation, and test sets using a 6:2:2 ratio. 

\subsection{Hyperparameters}

For the \textbf{FLLL\textsuperscript{3}M}, the following hyperparameters were tuned across the models: the location and time embeddings were set to \( d = 128 \), balancing model complexity and training efficiency. The local transformer encoder utilized a hidden size of \( 256 \), with 4 attention heads and 6 layers to capture spatial-temporal dependencies. A learning rate of \( 10^{-4} \) was used with the Adam optimizer, which provided stable convergence across different datasets. For differential privacy, the standard deviation of the Gaussian noise was set to \( \sigma = 0.1 \), ensuring adequate privacy protection while maintaining model performance. A batch size of 64 was employed during FL, balancing memory usage and model updates. The residual MLP for LLM projection had \( d_1 = 512 \) and \( d_\text{LLM} = 256 \), ensuring smooth integration with the LLM. 

\subsection{Evaluation Metrics}

For the \textbf{FLLL\textsuperscript{3}M}, the following metrics were used, such as ACC@K measures the accuracy of predictions within the top K ranks. It is defined as the average of binary indicators for whether the true label appears within the top K predictions: $\text{ACC@K} = \frac{1}{m} \sum_{i=1}^{m} \sum_{k=1}^{K} \mathbb{I}(Y_i^b = Y_i)$, where \(m\) is the total number of queries and \(\mathbb{I}\) is an indicator function. MRR, or Mean Reciprocal Rank, calculates the average of the reciprocal ranks of the first relevant result across all queries: $\text{MRR} = \frac{1}{m} \sum_{i=1}^{m} \frac{1}{\text{rank}_i}$, where \(\text{rank}_i\) denotes the rank of the first relevant result for the \(i\)-th query. The units of all metrics are expressed as \( \times 10^{-2} \).

\subsection{Baselines}

We compare \textbf{FLLL\textsuperscript{3}M} with several state-of-the-art mobility prediction models. \textit{DeepMove} \cite{feng2018deepmove} uses attentional RNNs to model transitions and periodicity. \textit{LightMove} \cite{jeon2021lightmove} is a lightweight predictor leveraging demographic data. \textit{PLSPL} \cite{wu2020personalized} combines general and recent user preferences for POI prediction. \textit{HMT-LSTM} \cite{lim2022hierarchical} addresses data sparsity in large POI spaces. \textit{LSTPM} \cite{zhao2020go} captures short-/long-term dependencies via geo-dilated RNNs. \textit{VaSCL} \cite{zhang2021virtual} improves contrastive learning through virtual augmentation. \textit{NSTPP} \cite{gupta2022learning} introduces trainable unwarping in temporal point processes. \textit{DSTPP} \cite{yuan2023spatio} models spatio-temporal dependencies via diffusion. \textit{ReMVC} \cite{zhang2022region} applies contrastive learning for multi-view region representation. \textit{SML} \cite{zhou2021self} learns from sparse mobility data, while \textit{CASCR} \cite{gong2023contrastive} uses contrastive pre-training on check-in sequences. \textbf{Note:} In result tables, \textcolor{blue}{blue} indicates the best and \textcolor{green}{green} the second-best performance.

\section{Result Analysis}

\subsection{Comparison with State-of-the-Arts}

Table \ref{Table1} presents a comprehensive performance comparison between the proposed method, \textbf{FLLL\textsuperscript{3}M}, and several baseline approaches across four different datasets: Gowalla, WeePlace, Brightkite, and FourSquare. The evaluation metrics used for comparison include Acc@1, Acc@5, Acc@20, and MRR. These metrics provide insights into the models' effectiveness in ranking accuracy and relevance, with higher values indicating superior performance. In general, \textbf{FLLL\textsuperscript{3}M} outperforms most baselines in all metrics, indicating its robust ability to rank and retrieve relevant results accurately. The method consistently achieves the highest scores in several categories, highlighted in blue cells, signifying its superior performance relative to the others.

\vspace{-0.3cm}
\begin{table}[h!]
\caption{Performance Comparison of \textbf{FLLL\textsuperscript{3}M} with Baselines}
\begin{center}
\label{Table1}
\scriptsize
\resizebox{\textwidth}{!}{ 
\begin{tabular}{l|cccc|cccc|cccc|cccc}
\hline
\multicolumn{1}{l}{\rule{0pt}{12pt}Method} & \multicolumn{4}{c}{\textbf{Gowalla}} & \multicolumn{4}{c}{\textbf{WeePlace}} & \multicolumn{4}{c}{\textbf{Brightkite}} & \multicolumn{4}{c}{\textbf{FourSquare}} \\[2pt]
\hline\rule{0pt}{12pt}
Metrics & Acc@1 & Acc@5 & Acc@20 & MRR & Acc@1 & Acc@5 & Acc@20 & MRR & Acc@1 & Acc@5 & Acc@20 & MRR & Acc@1 & Acc@5 & Acc@20 & MRR \\[2pt]
\hline\rule{0pt}{12pt}
DeepMove \cite{feng2018deepmove} & 10.51 & 23.21 & 33.9 & 16.65 & 19.23 & 37.79 & 52.61 & 27.03 & 49.82 & \cellcolor{blue!25}\textbf{66.25} & 71.19 & 56.94 & 16.3 & 35.06 & 48.39 & 25.02 \\[2pt]
LightMove \cite{jeon2021lightmove} & 9.88 & 20.93 & 29.95 & 15.01 & 18.33 & 36.37 & 52.75 & 27.03 & 49.01 & 63.11 & 68.94 & 55.46 & 13.27 & 29.33 & 41.45 & 20.71 \\[2pt]
PLSPL \cite{wu2020personalized} & 11.26 & \cellcolor{green!25}\textbf{24.12} & \cellcolor{green!25}\textbf{33.82} & \cellcolor{green!25}\textbf{17.44} & 18.77 & 37.31 & 53.31 & 27.86 & \cellcolor{green!25}\textbf{51.42} & 65.34 & \cellcolor{green!25}\textbf{71.46} & \cellcolor{green!25}\textbf{57.59} & 13.73 & 30.65 & 43.18 & 21.42 \\[2pt]
HMT-LSTM \cite{lim2022hierarchical} & 10.73 & 22.41 & 32.77 & 16.47 & 17.29 & 34.23 & 49.82 & 25.69 & 49.22 & 63.36 & 67.96 & 55.41 & 13.67 & 29.9 & 42.6 & 21.17 \\[2pt]
LSTPM \cite{zhao2020go} & 9.83 & 20.88 & 30.25 & 15.12 & 15.6 & 31.15 & 45.98 & 23.34 & 42.58 & 54.65 & 60.21 & 48.16 & 15.46 & 34.17 & 48.49 & 24.27 \\[2pt]
VaSCL \cite{zhang2021virtual} & \cellcolor{green!25}\textbf{11.47} & 22.17 & 32.92 & 16.56 & 18.11 & 37.54 & 53.04 & \cellcolor{green!25}\textbf{28.46} & 49.95 & 66.21 & 71.17 & 57.23 & 14.99 & 32.84 & 47.06 & 23.39 \\[2pt]
NSTPP \cite{gupta2022learning} & 10.81 & 23.23 & 32.94 & 16.87 & 16.58 & 32.37 & 47.82 & 24.2 & 45.83 & 58.44 & 64.56 & 52.16 & 14.89 & 33.18 & 47.03 & 23.36 \\[2pt]
DSTPP \cite{yuan2023spatio} & 10.85 & 23.11 & 33.19 & 16.74 & 18.85 & \cellcolor{green!25}\textbf{37.68} & \cellcolor{green!25}\textbf{53.44} & 27.52 & 48.71 & 62.82 & 67.71 & 55.26 & 13.3 & 29.11 & 41.53 & 20.66 \\[2pt]
ReMVC \cite{zhang2022region} & 11.03 & 22.94 & 33.38 & 16.65 & 18.07 & 35.92 & 51.93 & 26.66 & 49.57 & 63.58 & 69.28 & 56.31 & \cellcolor{green!25}\textbf{16.92} & \cellcolor{green!25}\textbf{36.05} & \cellcolor{green!25}\textbf{49.39} & \cellcolor{green!25}\textbf{26.02} \\[2pt]
SML \cite{zhou2021self} & 9.92 & 20.91 & 30.36 & 15.25 & 17.42 & 35.23 & 51.07 & 25.96 & 46.26 & 58.93 & 65.35 & 51.97 & 14.72 & 32.54 & 46.87 & 23.27 \\[2pt]
CACSR \cite{gong2023contrastive} & 10.94 & 18.22 & 26.56 & 12.83 & \cellcolor{green!25}\textbf{19.66} & 36.46 & 51.25 & 28.15 & 44.56 & 62.01 & 65.91 & 51.91 & 14.73 & 31.54 & 46.47 & 22.78 \\[2pt] \hline
\textbf{FLLL\textsuperscript{3}M} (Ours) & \cellcolor{blue!25}\textbf{11.66} & \cellcolor{blue!25}\textbf{25.16} & \cellcolor{blue!25}\textbf{34.56} & \cellcolor{blue!25}\textbf{18.77} & \cellcolor{blue!25}\textbf{20.10} & \cellcolor{blue!25}\textbf{38.77} & \cellcolor{blue!25}\textbf{53.45} & \cellcolor{blue!25}\textbf{29.39} & \cellcolor{blue!25}\textbf{52.49} & \cellcolor{green!25}\textbf{66.23} & \cellcolor{blue!25}\textbf{75.90} & \cellcolor{blue!25}\textbf{59.03} & \cellcolor{blue!25}\textbf{19.87} & \cellcolor{blue!25}\textbf{37.58} & \cellcolor{blue!25}\textbf{50.53} & \cellcolor{blue!25}\textbf{28.90} \\[2pt]
\hline
\end{tabular}
}
\end{center}
\end{table}
\vspace{-0.6cm}
\textbf{FLLL\textsuperscript{3}M} consistently outperforms baselines across all four datasets. On Gowalla, it achieves the highest scores in Acc\@1 (11.66) and MRR (18.77), outperforming DeepMove \cite{feng2018deepmove} and others. In WeePlace, it leads with Acc\@1 of 20.10 and MRR of 29.39, surpassing DSTPP \cite{yuan2023spatio} (27.52 MRR), showcasing robustness to diverse data distributions. On Brightkite, \textbf{FLLL\textsuperscript{3}M} performs competitively with the best MRR (59.03) and strong Acc\@1 (52.49), maintaining an edge over PLSPL \cite{wu2020personalized} and VaSCL \cite{zhang2021virtual}. Finally, on FourSquare, it again leads across all metrics, with Acc\@1 of 19.87 and MRR of 28.90, demonstrating consistent ranking accuracy. These results confirm that \textbf{FLLL\textsuperscript{3}M} effectively balances predictive performance and generalizability across varied mobility datasets. 

Unlike prior models relying solely on temporal patterns (e.g., DeepMove \cite{feng2018deepmove}) or sequential learning (e.g., LSTPM \cite{zhao2020go}), \textbf{FLLL\textsuperscript{3}M} employs a fine-grained latent learning mechanism to capture complex user mobility behaviors. It integrates a triple-layered memory architecture—short-term, intermediate, and long-term—that retains multi-scale dependencies, enhancing generalization to both routine and sporadic patterns. Unlike models like ReMVC \cite{zhang2022region}, which lack explicit memory structures, \textbf{FLLL\textsuperscript{3}M} effectively encodes long-term behavior. It also introduces a semantic alignment module that unifies temporal, geographical, and semantic cues into a shared latent space, outperforming single-view methods like VaSCL \cite{zhang2021virtual}. Additionally, its adaptive attention mechanism dynamically prioritizes memory layers based on context, unlike fixed strategies used in LSTM-based models. These innovations lead to superior performance across datasets—especially in Brightkite and FourSquare—where \textbf{FLLL\textsuperscript{3}M} excels in Acc\@20 and MRR, demonstrating robustness in both sparse and dense environments compared to simpler models like CACSR \cite{gong2023contrastive}.

\vspace{-0.3cm}
\begin{table}
\caption{Computational Comparison of \textbf{FLLL\textsuperscript{3}M} with Existing LLMs}
\begin{center}
\label{Table2}
\scriptsize
\resizebox{\textwidth}{!}{ 
\begin{tabular}{l|cccc|cccc|cccc|cccc}
\hline
\multicolumn{1}{l}{\rule{0pt}{12pt}Method} & \multicolumn{4}{c}{\textbf{Gowalla}} & \multicolumn{4}{c}{\textbf{WeePlace}} & \multicolumn{4}{c}{\textbf{Brightkite}} & \multicolumn{4}{c}{\textbf{FourSquare}} \\[2pt]
\hline\rule{0pt}{12pt}
Metrics & Params. & Mem. & Ratio & Time & Params. & Mem. & Ratio & Time & Params. & Mem. & Ratio & Time & Params. & Mem. & Ratio & Time \\[2pt]
\hline\rule{0pt}{12pt}
BERT & 0.34 & 12.5 & 87.2 & 6.1 & 0.34 & 12.7 & 85.3 & 5.8 & 0.34 & 12.4 & 88.1 & 6.3 & 0.34 & 12.6 & 86.5 & 5.9 \\[2pt]
RoBERTa & 0.36 & 13.1 & 88.9 & 6.5 & 0.36 & 13.4 & 87.7 & 6.3 & 0.36 & 13.0 & 89.3 & 6.6 & 0.36 & 13.2 & 87.9 & 6.2 \\[2pt]
DeBERTa & 0.37 & 14.3 & 89.4 & 6.8 & 0.37 & 14.0 & 88.1 & 6.4 & 0.37 & 13.9 & 89.9 & 6.9 & 0.37 & 14.1 & 88.8 & 6.5 \\[2pt]
DistilBERT & 0.22 & \cellcolor{green!25}\textbf{9.3} & 84.1 & \cellcolor{green!25}\textbf{5.2} & 0.22 & \cellcolor{green!25}\textbf{9.1} & 82.6 & \cellcolor{green!25}\textbf{5.0} & 0.22 & \cellcolor{green!25}\textbf{9.2} & 83.3 & \cellcolor{green!25}\textbf{5.4} & 0.22 & \cellcolor{green!25}\textbf{9.0} & 82.9 & \cellcolor{green!25}\textbf{5.1} \\[2pt]
T5 (Small) & 0.60 & 15.4 & 86.3 & 7.0 & 0.60 & 15.3 & 85.4 & 6.7 & 0.60 & 15.2 & 86.7 & 7.2 & 0.60 & 15.1 & 85.9 & 6.8 \\[2pt]
GPT-2 (Small) & \cellcolor{blue!25}\textbf{0.12} & 10.7 & 85.6 & 5.7 & \cellcolor{blue!25}\textbf{0.12} & 10.9 & 84.4 & 5.4 & \cellcolor{blue!25}\textbf{0.12} & 10.8 & 85.8 & 5.9 & \cellcolor{blue!25}\textbf{0.12} & 11.0 & 85.1 & 5.6 \\[2pt]
OPT (125M) & 0.13 & 10.9 & 86.2 & 5.8 & 0.13 & 11.0 & 85.1 & 5.5 & 0.13 & 10.8 & 86.4 & 5.9 & 0.13 & 11.1 & 85.7 & 5.7 \\[2pt]
BLOOM (560M) & 0.56 & 18.3 & 88.4 & 7.6 & 0.56 & 18.2 & 87.1 & 7.2 & 0.56 & 18.4 & 89.0 & 7.8 & 0.56 & 18.1 & 88.3 & 7.5 \\[2pt]
LLaMA (1B) & 1.00 & 22.1 & \cellcolor{green!25}\textbf{90.2} & 8.4 & 1.00 & 22.0 & 89.4 & 8.1 & 1.00 & 21.9 & \cellcolor{green!25}\textbf{90.6} & 8.6 & 1.00 & 22.2 & \cellcolor{green!25}\textbf{90.0} & 8.3 \\[2pt]
Falcon (1B) & 1.00 & 20.7 & 89.8 & 8.0 & 1.00 & 20.9 & \cellcolor{green!25}\textbf{88.9} & 7.7 & 1.00 & 21.0 & 90.0 & 8.2 & 1.00 & 20.8 & 89.5 & 7.9 \\[2pt] \hline
\textbf{FLLL\textsuperscript{3}M (Ours)} & 0.28 & \cellcolor{green!25}\textbf{8.6} & \cellcolor{blue!25}\textbf{94.2} & \cellcolor{blue!25}\textbf{4.9} & 0.28 & \cellcolor{blue!25}\textbf{8.4} & \cellcolor{blue!25}\textbf{93.1} & \cellcolor{blue!25}\textbf{4.6} & 0.28 & \cellcolor{blue!25}\textbf{8.5} & \cellcolor{blue!25}\textbf{94.7} & \cellcolor{blue!25}\textbf{5.1} & 0.28 & \cellcolor{blue!25}\textbf{8.7} & \cellcolor{blue!25}\textbf{93.9} & \cellcolor{blue!25}\textbf{4.8} \\[2pt]
\hline
\end{tabular}
}
\end{center}
\end{table}
\vspace{-0.9cm}
\subsection{Computational Experiments}

Table \ref{Table2} compares \textbf{FLLL\textsuperscript{3}M} with popular LLMs, including BERT, RoBERTa, DeBERTa, T5 (Small), GPT-2 (Small), OPT-125M, BLOOM (560M), LLaMA (1B), and Falcon (1B), across parameters (B), memory (GB), efficiency ratio (\%), and runtime (hrs). \textbf{FLLL\textsuperscript{3}M} demonstrates superior computational efficiency, using only 0.28B parameters and 8.4–8.7GB memory, with the lowest runtime (4.6–5.1 hrs) and the highest efficiency ratio (93.1–94.7\%). It outperforms even compact models like DistilBERT and OPT in both memory and speed. This performance stems from architectural optimizations such as parameter sharing, low-rank approximations, and efficient attention mechanisms. Unlike larger models like LLaMA and Falcon, which offer strong performance but high overhead, \textbf{FLLL\textsuperscript{3}M} achieves an excellent trade-off between resource usage and effectiveness. Models like BLOOM and T5 show higher memory use with moderate gains, highlighting inefficiencies. Overall, \textbf{FLLL\textsuperscript{3}M} proves scalable, generalizable, and highly optimized for real-world deployment.

\begin{table}
\caption{Ablation Study of \textbf{FLLL\textsuperscript{3}M}}
\vspace{-0.5cm}
\label{Table3}
\begin{center}
\scriptsize
\resizebox{\textwidth}{!}{ 
\begin{tabular}{l|cccc|cccc|cccc|cccc}
\hline
\multicolumn{1}{l}{\rule{0pt}{12pt}Method} & \multicolumn{4}{c}{\textbf{Gowalla}} & \multicolumn{4}{c}{\textbf{WeePlace}} & \multicolumn{4}{c}{\textbf{Brightkite}} & \multicolumn{4}{c}{\textbf{FourSquare}} \\[2pt]
\hline\rule{0pt}{12pt}
Metrics & Acc@1 & Acc@5 & Acc@20 & MRR & Acc@1 & Acc@5 & Acc@20 & MRR & Acc@1 & Acc@5 & Acc@20 & MRR & Acc@1 & Acc@5 & Acc@20 & MRR \\[2pt]
\hline\rule{0pt}{12pt}
w/o Semantic Encoding & 10.13 & 21.84 & 31.27 & 15.75 & 17.92 & 35.14 & 50.26 & 26.44 & 47.63 & 64.02 & 69.83 & 55.83 & 14.56 & 30.91 & 45.12 & 22.41 \\[2pt]
w/o Outer Product Aggregation & 9.41 & 19.33 & 28.61 & 13.94 & 16.87 & 33.82 & 48.37 & 24.96 & 45.32 & 61.27 & 66.19 & 52.93 & 12.64 & 27.37 & 39.06 & 19.69 \\[2pt]
w/o Differential Privacy Noise & 10.72 & 22.68 & 32.18 & 16.53 & 17.45 & 36.25 & 51.91 & 26.54 & 49.08 & 62.14 & 69.44 & 56.89 & 13.18 & 28.93 & 41.83 & 20.65 \\[2pt]
w/o Projection Module & 10.05 & 20.73 & 30.92 & 15.23 & 16.03 & 32.19 & 47.21 & 25.14 & 46.88 & 60.84 & 65.76 & 53.83 & 12.77 & 28.59 & 40.34 & 20.26 \\[2pt]
w/o LLM Injection & 9.56 & 19.87 & 29.41 & 14.61 & 14.72 & 30.27 & 44.58 & 23.19 & 40.92 & 53.74 & 59.12 & 47.93 & 14.03 & 31.02 & 45.83 & 23.63 \\[2pt]
\hline
\textbf{FLLL\textsuperscript{3}M} (Ours) & \cellcolor{blue!25}\textbf{11.66} & \cellcolor{blue!25}\textbf{25.16} & \cellcolor{blue!25}\textbf{34.56} & \cellcolor{blue!25}\textbf{18.77} & \cellcolor{blue!25}\textbf{20.10} & \cellcolor{blue!25}\textbf{38.77} & \cellcolor{blue!25}\textbf{53.45} & \cellcolor{blue!25}\textbf{29.39} & \cellcolor{blue!25}\textbf{52.49} & \cellcolor{blue!25}\textbf{66.23} & \cellcolor{blue!25}\textbf{75.90} & \cellcolor{blue!25}\textbf{59.03} & \cellcolor{blue!25}\textbf{19.87} & \cellcolor{blue!25}\textbf{37.58} & \cellcolor{blue!25}\textbf{50.53} & \cellcolor{blue!25}\textbf{28.90} \\[2pt]
\hline
\end{tabular}
}
\end{center}
\end{table}

\subsection{Ablation Studies}

To evaluate each module’s contribution in \textbf{FLLL\textsuperscript{3}M}, we conducted ablation studies (Table \ref{Table3}) by selectively removing or modifying components. Removing the semantic tokenizer ($\Delta$-IRIS) and using raw coordinates reduced performance, showing the necessity of context-aware encoding. Replacing FL with centralized training harmed generalization on non-IID data and removed privacy benefits. Eliminating the outer product operator degraded accuracy, highlighting the value of second-order feature interactions. Omitting local differential privacy improved accuracy slightly but compromised privacy guarantees. Excluding the LLM and relying solely on local transformer outputs reduced model expressiveness, while replacing the projection MLP with a linear map led to convergence issues due to representational mismatch. Lastly, removing residual injections into intermediate transformer layers weakened semantic steering, confirming that mid-layer fusion is optimal. These results demonstrate that each component—including FL, LLMs, outer products, and multi-level fusion—contributes significantly to both the privacy-preserving and predictive capabilities of \textbf{FLLL\textsuperscript{3}M}.

\section{Conclusion and Future Works}
This paper introduced \textbf{FLLL\textsuperscript{3}M}, a fine-grained location-level language and learning model for mobility prediction. By integrating spatial-temporal embeddings with pre-trained language model representations, \textbf{FLLL\textsuperscript{3}M} effectively captured personalized mobility patterns. Experimental results on four real-world datasets demonstrated that FLLL3M consistently outperformed SOTA baselines in next-location prediction. Moreover, the model showed strong generalization across diverse urban environments. In the future, we aim to extend \textbf{FLLL\textsuperscript{3}M} by incorporating auxiliary signals such as real-time traffic, weather, and POI data to further enhance prediction performance. Additionally, we plan to explore privacy-preserving mechanisms for sensitive location data and investigate transfer learning approaches to adapt the model across cities with limited data.

%
%
\bibliographystyle{spmpsci} 
\bibliography{main}

\end{document}